\documentclass{article}
\usepackage{amsmath,amssymb,amsfonts}
\usepackage{arxiv}
\usepackage{algorithm}
\usepackage{algorithmic}
\usepackage{algorithmic}
\newcommand{\Comment}[1]{\hfill\textit{#1}}
\usepackage[utf8]{inputenc} % allow utf-8 input
\usepackage[T1]{fontenc}    % use 8-bit T1 fonts
\usepackage{hyperref}       % hyperlinks
\usepackage{url}            % simple URL typesetting
\usepackage{booktabs}       % professional-quality tables       % blackboard math symbols
\usepackage{nicefrac}       % compact symbols for 1/2, etc.
\usepackage{microtype}      % microtypography
\usepackage{lipsum}		% Can be removed after putting your text content
\usepackage[numbers]{natbib}
\usepackage{doi}
\usepackage{graphicx}   % For including graphics
\usepackage{caption}    % For customizing captions
\usepackage{cleveref}   % For improved cross-referencing
\usepackage{array}
\usepackage{makecell}
\usepackage{tabularx}
\usepackage{adjustbox}
\usepackage{float}
\usepackage{enumitem}
\usepackage{longtable}
\usepackage{geometry}
\newcommand{\figref}[1]{\hyperref[#1]{Figure\ref*{#1}}}

\newcommand{\tabref}[1]{\hyperref[#1]{Table\ref*{#1}}}

\newcommand{\Algref}[1]{\hyperref[#1]{Algorithm\ref*{#1}}}
\renewcommand{\arraystretch}{1.3} % Adjust the row height
\title{A Novel Multi-Task Teacher–Student Architecture with Self-Supervised Pretraining for 48-Hour Vasoactive-Inotropic Trend Analysis in Sepsis Mortality Prediction}

\author{ 
    \textbf{Houji Jin}$^{1}$, \textbf{Negin Ashrafi}$^{1}$, \textbf{Kamiar Alaei}$^{2}$, \\
    \textbf{Elham Pishgar}$^{3}$, \textbf{Greg Placencia}$^{4}$, \textbf{Maryam Pishgar}$^{1*}$ \\
    \\
$^{1}$ University of Southern California, Los Angeles, CA, USA \\
$^{2}$ California State University, Long Beach, CA, USA \\
$^{3}$ Iran University of Medical Sciences, Tehran, Iran \\
$^{4}$ California State Polytechnic University, Pomona, CA, USA \\
    *Corresponding author: \texttt{pishgar@usc.edu}
}

\renewcommand{\undertitle}{Research Paper}

\begin{document}
\maketitle
\begin{abstract}
Sepsis is a major cause of ICU mortality, where early recognition and effective interventions are essential for improving patient outcomes. However, the vasoactive-inotropic score (VIS) varies dynamically with a patient's hemodynamic status, complicated by irregular medication patterns, missing data, and confounders, making sepsis prediction challenging. To address this, we propose a novel Teacher–Student multitask framework with self-supervised VIS pretraining via a Masked Autoencoder (MAE). The teacher model performs mortality classification and severity-score regression, while the student distills robust time-series representations, enhancing adaptation to heterogeneous VIS data. Compared to LSTM-based methods, our approach achieves an AUROC of 0.82 on MIMIC-IV 3.0 (9,476 patients), outperforming the baseline (0.74). SHAP analysis revealed that SOFA score (0.147) had the greatest impact on ICU mortality, followed by LODS (0.033), single marital status (0.031), and Medicaid insurance (0.023), highlighting the role of sociodemographic factors. SAPSII (0.020) also contributed significantly. These findings suggest that both clinical and social factors should be considered in ICU decision-making. Our novel multitask and distillation strategies enable earlier identification of high-risk patients, improving prediction accuracy and disease management, offering new tools for ICU decision support.

\end{abstract}


\maketitle

\section{Introduction}
Sepsis is a syndrome that results from a dysregulated response to infection leading to life-threatening organ dysfunction \cite{Singer2016, Cecconi2018}. It remains one of the primary causes of morbidity and mortality in intensive care units (ICUs) \cite{Singer2016}. Clinically, stabilizing the hemodynamic status of patients with septic shock requires the administration of a vasoactive–inotropic score (VIS), which includes medications such as dopamine, norepinephrine, dobutamine, epinephrine, milrinone, and vasopressin. However, accurately modeling and predicting the dosage and trends of these medications within the first 48 hours is challenging due to the variable physiology of ICU patients, irregular and incomplete medication records, and the presence of numerous clinical confounding factors \cite{Johnson2024MIMIC, Johnson2023SciData, Goldberger2000}.

Traditional ICU prognostic tools (e.g., models based on logistic regression or single scores) provide some early risk assessment but often fail to adequately capture time-series dynamics, struggle with high-dimensional heterogeneous data, and are limited in both interpretability and generalizability \cite{Komorowski2018}. In response to these challenges, recent machine learning–based methods have been developed to enhance sepsis mortality prediction and early detection. For instance, Gao et al. \cite{Gao2024} implemented a model leveraging a carefully selected set of 38 features from the MIMIC-IV database, successfully balancing predictive accuracy with enhanced interpretability. Yu et al. \cite{Yu2024} focused on predicting 30-day mortality for ICU patients with Sepsis-3 by extracting 30 critical features, demonstrating the power of decision-tree frameworks and big data processing tools for robust risk stratification. Building on these approaches, Shumilov et al. \cite{Shumilov2024} proposed a data-driven machine learning framework incorporating random forest and advanced feature selection to achieve state-of-the-art performance, notably an AUROC of 0.97, in predicting in-hospital mortality among MIMIC-III patients. Taken together, these studies underscore the importance of clinically informed machine learning frameworks for improving sepsis outcomes in modern ICUs.

With the rapid advancement of deep learning in healthcare, researchers have employed recurrent neural networks (RNNs), long-short-term memory networks (LSTMs) \cite{Ning2023}
, and Transformer-based architectures to model ICU time-series data with greater precision. However, these models remain susceptible to overfitting and unstable performance when dealing with missing \cite{Shashikumar2021}, noisy, or irregularly sampled data in real ICU environments. Moreover, clinical practice frequently involves multiple correlated tasks, such as mortality prediction in conjunction with assessing organ function or disease severity scores, necessitating models that can effectively share latent representations across different tasks and mitigate bias from over-reliance on a single objective \cite{Ruder2017, Sener2018}. Advances in optimization, including adaptive momentum and decentralized learning, have improved model stability and scalability in clinical time-series analysis \cite{Barazandeh2021, Barazandeh2021RSRC, Barazandeh2021Decentralized}."

In recent years, the Masked Autoencoder (MAE) approach has made significant strides in computer vision and Natural Language Processing (NLP) \cite{He2022, Devlin2018}. This method learns robust feature representations in an unsupervised or self-supervised manner by masking parts of the input and training the network to reconstruct them. Similarly, MAE offers a natural advantage for addressing the frequent issues of missing or irregular sampling in ICU time-series data, as it enhances the network’s resilience to incomplete or noisy inputs through its "masking" training process \cite{He2022, Desautels2016}. Generative augmentation techniques, such as variational autoencoders and generative adversarial networks, have also been shown to enhance robustness and interpretability in medical diagnosis tasks \cite{Saadatinia2024}.

Meanwhile, knowledge distillation (KD) is increasingly attracting attention in the medical field \cite{Meng2021}. In its classical form, KD leverages a large-scale Teacher model to guide the training of a smaller-scale Student model, thereby reducing model complexity or inference cost while preserving performance. Historically, KD has been applied in areas such as computer vision and large language models \cite{Hinton2015, Gou2021, Gu2023}. For instance, Termritthikun, Chakkrit, et al. \cite{Termritthikun2023} employed KD to efficiently detect infectious regions in chest X-rays (COVID-19), reducing inference latency via a Teacher–Student architecture. Rohanian, Omid, et al. \cite{Rohanian2024} introduced TinyClinicalBERT, capable of performing Named Entity Recognition, Relation Extraction, and Sequence Classification, which significantly reduces the number of parameters while maintaining accuracy. Goldblum, Micah, et al. \cite{Goldblum2020} developed RobustKD, accelerating knowledge distillation by integrating adversarial training to enhance the Student model’s robustness under challenging data distributions. Moreover, Wong, Anna, et al. \cite{Wong2023} incorporated interpretability into the distillation process for early sepsis identification, underscoring the value of interpretable Teacher–Student methods in critical decision-making. Notably, in high-risk settings such as ICUs, deploying overly large models in real time is often impractical due to hardware constraints. The KD approach mitigates this issue by producing more lightweight models that retain accuracy and exhibit improved robustness to small or incomplete datasets. Reinforcement learning-based early warning systems have also demonstrated strong potential for improving timely sepsis detection in ICU settings \cite{Dai2023}."

Finally, Multi-Task Learning (MTL) represents another branch of deep learning that enhances overall learning efficiency and generalization by sharing parameters across related objectives \cite{Ruder2017, Sener2018}. In sepsis management, beyond predicting a patient's risk of death, it is critical to assess organ function or disease severity (e.g., SOFA, LODS, SAPSII) \cite{Knaus1985}. By training a single model for both mortality classification and severity score regression, a richer representation of temporal features can be achieved, thereby improving both the interpretability and robustness of the model.

Based on these motivations, we propose a novel "teacher–student" multitask architecture for predicting ICU mortality in septic patients, utilizing 7-dimensional VIS features for Masked Autoencoder (MAE) pretraining. Initially, the teacher model is trained on a self-supervised MAE task and then concatenated with an untrained downstream multitask learning network (for mortality classification and severity score regression) to learn robust representations despite noisy and missing time-series data. Subsequently, the student model inherits and refines the teacher’s time-series representations through knowledge distillation. Our evaluation on the MIMIC-IV 3.0 database \cite{Johnson2024MIMIC} of 9,476 sepsis patients demonstrates that the AUROC can reach 0.82, a significant improvement over the previous LSTM-based model (0.74) \cite{Ning2023}. Furthermore, interpretability analysis using SHAP \cite{Lundberg2017} revealed that both clinical severity scores (e.g., SOFA, LODS) and patient demographic factors (e.g., marital status, insurance type) significantly impact mortality prediction. These findings underscore the need for a comprehensive, multifactorial approach to managing septic shock. Overall, this study presents a practical strategy to overcome challenges in previous ICU mortality prediction models, including difficulties in capturing irregular time-series data, lack of multitask synergy, and limited interpretability, ultimately aiming to enhance clinical decision-making and improve patient outcomes.

\section{Data Acquisition and Experimental Setup}
\label{sec:dataset_setup}

We utilized data from 9,476 sepsis patients (defined by Sepsis-3 criteria \cite{Singer2016}) in the MIMIC-IV v3.0 database \cite{Johnson2024MIMIC}. All patients received a vasoactive–inotropic score (VIS) within the first 48 hours of ICU admission, and those discharged against medical advice were excluded. Each patient’s 7-dimensional VIS time series included doses for dopamine, dobutamine, epinephrine, milrinone, vasopressin, norepinephrine, and total VIS. Additionally, static features—such as gender, admission age, insurance, marital status, and severity scores (SOFA, SAPS-II, LODS, OASIS)—were recorded, with ICU mortality serving as the binary outcome label. As shown in Equation~\eqref{eq:total_vis}, the VIS is calculated by summing the contributions of various medications.

\begin{equation}
\label{eq:total_vis}
\text{VIS} = \text{dopamine} + \text{dobutamine} + 100\,\text{epinephrine} + 10\,\text{milrinone} + 100\,\text{norepinephrine} + 10\,000\,\text{vasopressin}.
\end{equation}

In this formula, each medication dose is weighted according to its relative potency and clinical impact on hemodynamic support \cite{Song2021}. For instance, epinephrine and norepinephrine are multiplied by 100, while milrinone is multiplied by 10, and vasopressin by 10,000, reflecting their stronger vasoactive effects compared to dopamine and dobutamine \cite{Hajjar2013}. This weighted sum provides a comprehensive measure of the overall vasoactive support administered, which is critical for accurately assessing the severity of a patient’s condition in the ICU.

\begin{figure}[H]
    \centering
    \includegraphics[width=0.85\textwidth]{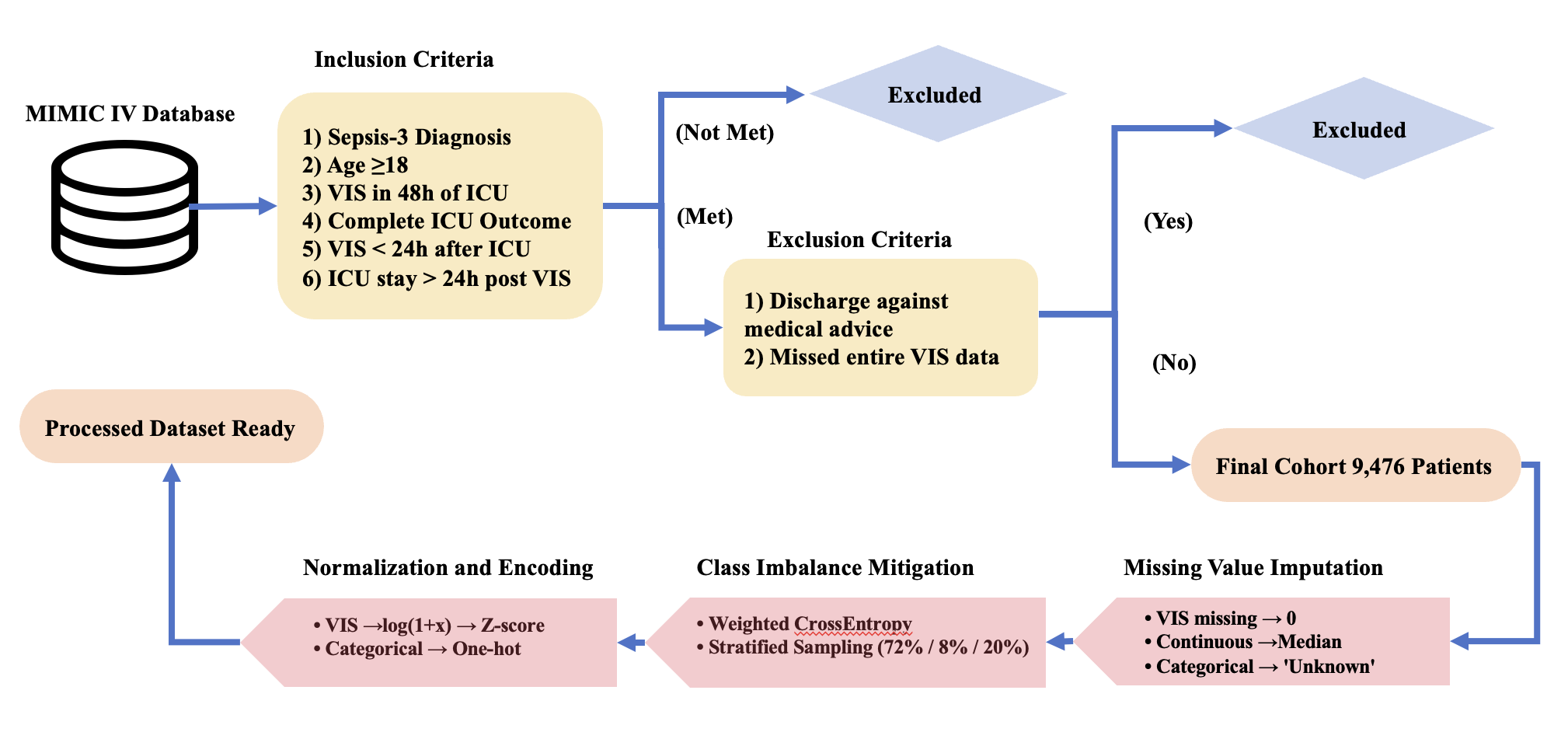} % Replace with your image filename
    \caption{Overview of the data extraction and preprocessing workflow.}
    \label{fig:flowchart}
\end{figure}

Below, we detail the background of the MIMIC-IV database, its applications in ICU research, and the data extraction and feature-engineering workflow. Our complete extraction and preprocessing process is illustrated in Figure~\ref{fig:flowchart}.

\subsection{Data Description}
The MIMIC-IV (Medical Information Mart for Intensive Care-IV) database, jointly developed by MIT and Beth Israel Deaconess Medical Center \cite{Johnson2024MIMIC}, is a publicly accessible repository of anonymized electronic health records that builds upon the foundations of MIMIC-III/II. It offers a comprehensive range of patient data—including demographics, diagnoses, laboratory tests, medications, nursing notes, vital signs, and outcomes—with timestamped, heterogeneous records that enable dynamic modeling of patient trajectories. This rich data composition is particularly valuable for sepsis research, as sepsis remains a leading cause of ICU mortality \cite{Singer2016, Goldberger2000} and often requires detailed analyses of vasoactive support and hemodynamic interventions. The availability of VIS dosing, laboratory values, and physiological measurements within MIMIC-IV facilitates in-depth investigations into sepsis progression and the effectiveness of clinical interventions. Overall, the database serves as an essential resource for developing advanced AI-driven clinical decision support systems, supporting rigorous and comprehensive analyses in the realm of critical care.

\subsection{Patient Inclusion Criteria}
Our cohort was constructed by applying specific inclusion and exclusion criteria. Patients were included if they had a Sepsis-3 diagnosis, were aged $\geq 18$ years, received VIS administration within the first 48 hours of ICU admission, had complete ICU outcome records (survival/death), and were administered VIS agents for therapy within 24 hours of ICU admission with an ICU stay lasting more than 24 hours following the commencement of vasoactive–inotropic therapy. Patients discharged against medical advice or those missing entire VIS infusion data were excluded. This resulted in a final cohort of 9,476 patients—slightly larger than the MIMIC-IV v2.0 cohort (8,887) in \cite{He2022}—reflecting the expanded coverage in version 3.0.

For missing value imputation, hourly missing doses in the VIS time series were imputed as 0 (indicating no drug use), while static continuous features (e.g., severity scores) were imputed using the median value. Categorical variables (e.g., marital status) incorporated an “Unknown” category for missing values.

To address the class imbalance between survival and death outcomes, we employed two strategies. First, a weighted cross-entropy loss was used, assigning higher weights to the minority class (death), with class weights computed as the total size of the training set divided by two times the class counts (see Equation~\eqref{eq:class_weights}). Second, stratified sampling was utilized to preserve the label distribution across training, validation, and test splits (0.72\%, 0.08\%, and 0.2\%, respectively).

\begin{equation}
\label{eq:class_weights}
\mathrm{class\_weights} = \frac{\mathrm{size}(\mathrm{train})}{2 \cdot \mathrm{class\_counts}}\,.
\end{equation}

The numerical features underwent a two-step normalization process. VIS values were first transformed using $\log(1+x)$ to suppress extreme values and stabilize the distribution, and then standardized via Z-score normalization using training-set parameters, which were also applied to the test and validation sets. Categorical features were one-hot encoded (e.g., gender, insurance). This preprocessing strategy facilitates model convergence and reduces the influence of extreme values, thus improving the model's ability to learn robust feature–label relationships.

\subsection{Feature Extraction}
We integrated three categories of features for multitask modeling. First, the VIS time series (7-dimensional) includes individual agents—dopamine, norepinephrine, vasopressin, milrinone, epinephrine, and dobutamine—as well as the total VIS, a weighted sum that reflects cumulative vasoactive support \cite{He2022}. Second, static features comprise demographics (gender, age, marital status, insurance, and race), which capture socioeconomic factors linked to ICU outcomes, and severity scores (SOFA, SAPS-II, LODS, and OASIS) that serve as regression targets for multitask learning by quantifying multiorgan dysfunction \cite{Wei2023, Knaus1985, Desautels2016}. Third, our feature selection is driven by several considerations: evidence-based predictors such as VIS dynamics \cite{Ning2023} and severity scores; the novelty of incorporating socioeconomic variables to address understudied prognostic factors; and interpretability, facilitated by SHAP \cite{Lundberg2017} to quantify feature contributions.

This integrated framework offers distinct comparative advantages by combining 7-dimensional VIS trends, severity scores, and socioeconomic variables to capture early (48-hour) hemodynamic responses and to identify vulnerable subgroups through an expanded feature space. In particular, socioeconomic factors are emphasized through a careful selection rationale. Marital status, for example, has been shown in retrospective analyses \cite{Seymour2010} to correlate with delayed clinical visits, increased hospitalization costs, and higher mortality, likely due to its association with social support and healthcare access. Similarly, insurance type and race, as evidenced in studies on severe sepsis in U.S. children \cite{Mitchell2021}, influence resource access and survival outcomes. Finally, while age and gender are routinely recorded, their established clinical significance—older age being associated with higher mortality risk and gender potentially affecting immune responses—reinforces their inclusion in our feature set.

\section{Overall Architecture}
\label{sec:arch}
The Vasoactive-Inotropic Score (VIS) quantifies the intensity of medications required to maintain hemodynamic stability in patients with septic shock, providing an objective measure for assessing disease progression and treatment efficacy \cite{Ning2023}. However, VIS data recorded in the ICU are often characterized by irregular sampling and missing values, posing significant challenges for time series modeling. The Masked AutoEncoder (MAE) addresses these challenges by randomly masking portions of the input and reconstructing them, thereby learning more robust feature representations under self-supervised conditions and reducing the model’s sensitivity to noise and missing data \cite{He2022}.

Recent studies have extended the MAE framework to various medical time series applications. For instance, Yang et al. \cite{Yang2022} applied an MAE-based approach for ECG representation learning, reporting notable improvements in downstream tasks. Similarly, Merkelbach et al. \cite{Merkelbach2023} introduced a gated recurrent unit autoencoder for ICU time series that effectively identified distinct patient subgroups, highlighting the potential of autoencoders to extract latent clinical phenotypes. In another study, Patel et al. \cite{Patel2024} developed an event-based masked autoencoding technique (EMIT) for irregular ICU data, demonstrating that integrating vital sign streams and laboratory measurements can yield robust performance even under sparse or noisy conditions. These advancements underscore the promise of self-supervised masking strategies for complex and heterogeneous healthcare data and suggest that further exploration of MAE-driven approaches in sepsis mortality prediction is warranted.

Our method integrates MAE with VIS data to capture the dynamic relationships between vasoactive medications and patient hemodynamic states, leveraging self-supervision to manage sparse or unstructured time series data. This approach not only enhances representation learning in high-noise ICU environments but also provides a more robust framework for sepsis prognosis and other critical-care predictions.

Our method unfolds in three integrated phases. Initially, we pretrain the MAE encoder by reconstructing randomly masked VIS data, enabling it to learn robust embeddings for the 48-hour time series despite noise and missing values \cite{He2022}. Subsequently, we partially freeze this MAE encoder and augment it with additional heads for ICU mortality classification and severity score regression, forming a teacher model that remains fixed during the subsequent training stage. Finally, an equally sized student model is trained on labeled data for both binary classification and regression, with knowledge distillation \cite{Hinton2015} employed to align its categorical output distribution with that of the teacher. This process effectively transfers the teacher’s robust learned representations to the student model, enhancing performance in both tasks.

\begin{figure}[!ht]
    \centering
    \includegraphics[width=0.75\textwidth]{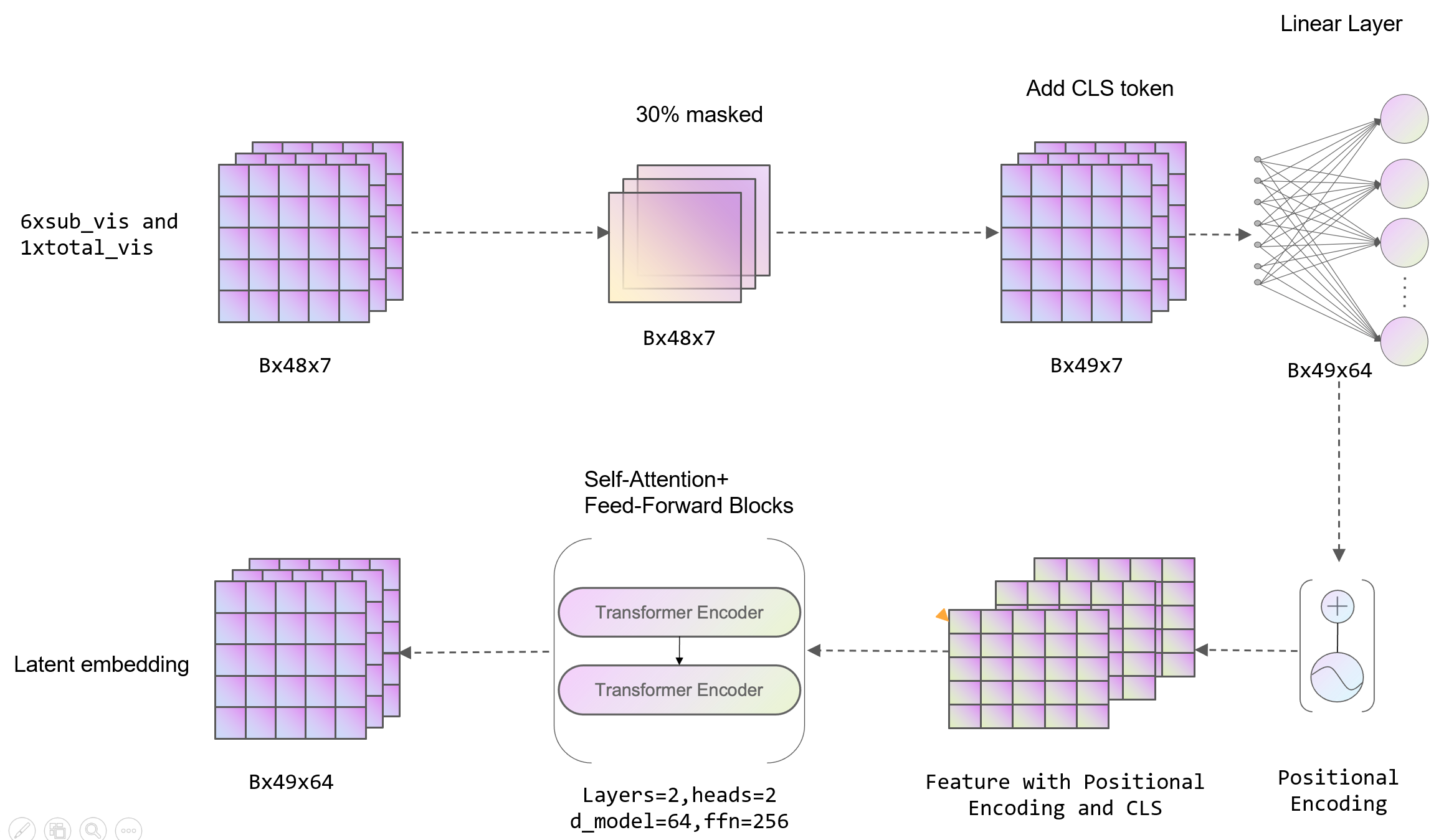}
    \caption{MAE\_Encoder architecture: shows token masking, [CLS] token prepending, linear projection to 64 dimensions, and a 2-layer Transformer (with $d_{\text{model}} = 64$, FFN = 256, 8 heads, 0.1 dropout) for VIS time series representation.}
    \label{fig:mae_enc}
\end{figure}

\begin{figure}[!ht]
    \centering
    \includegraphics[width=0.4\textwidth]{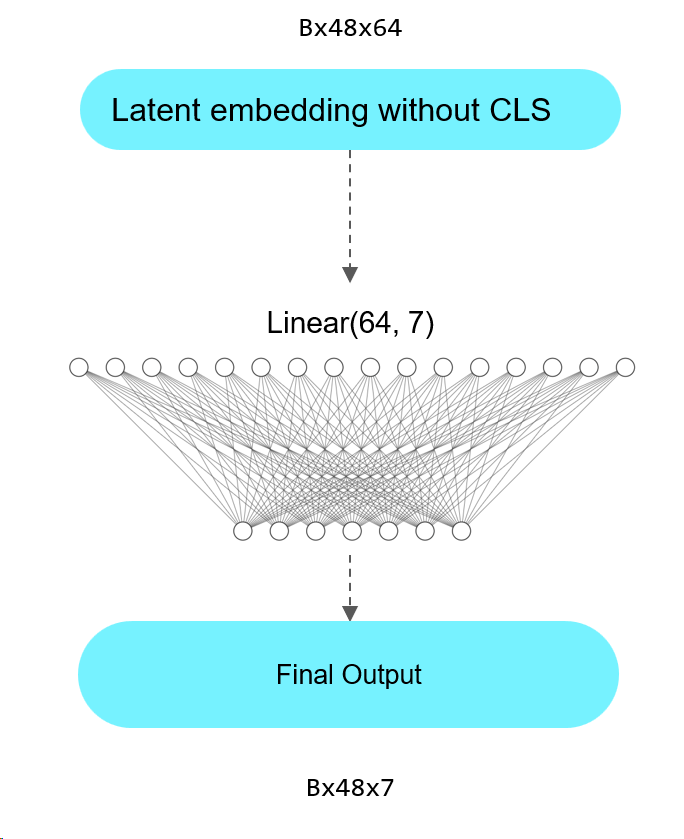}
    \caption{MAE\_Decoder schematic: reconstructs masked tokens from the latent embedding (excluding the [CLS] token) back to the original 7-dimensional VIS space.}
    \label{fig:mae_dec}
\end{figure}

\begin{figure}[!ht]
    \centering
    \includegraphics[width=0.9\textwidth]{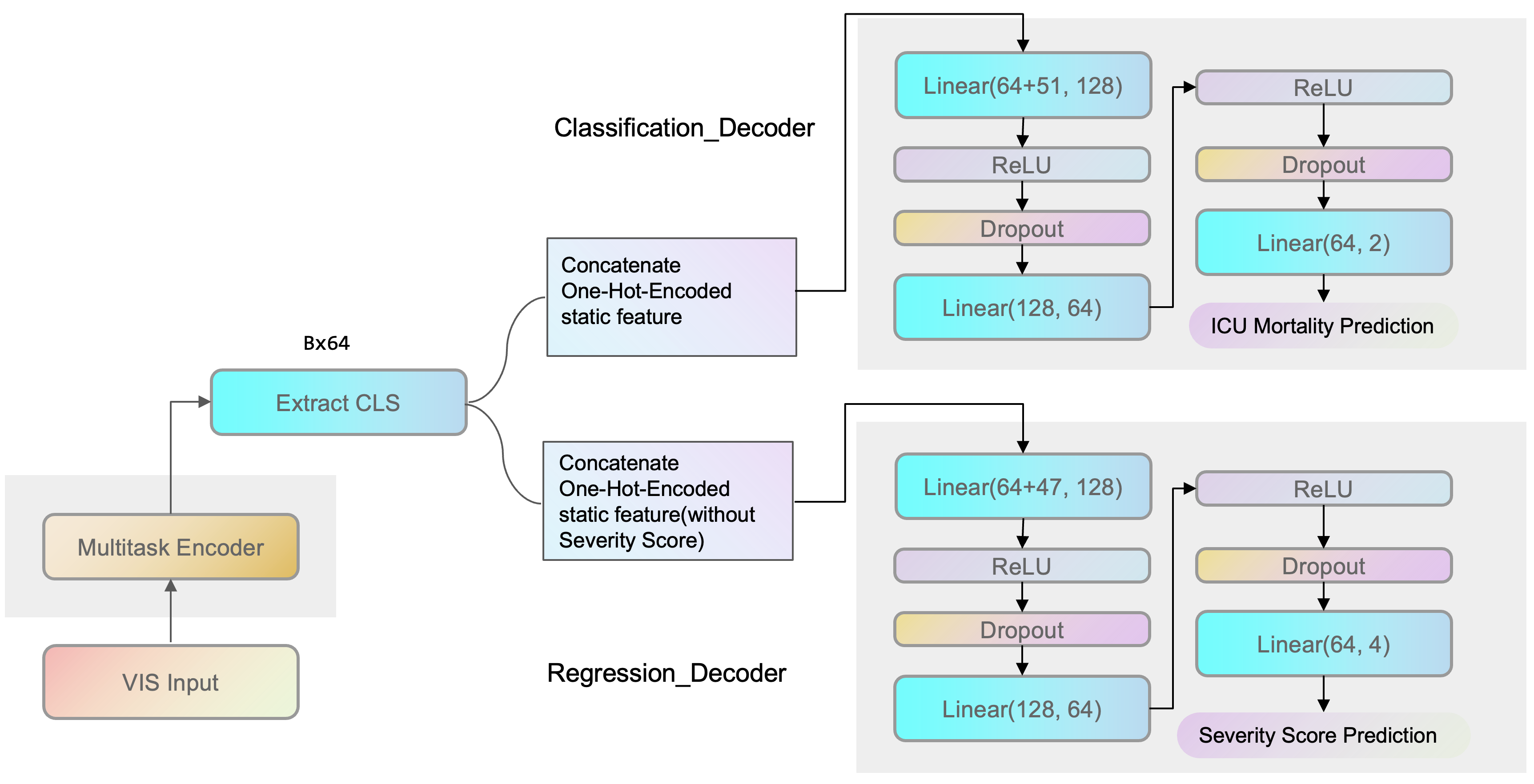}
    \caption{Multitask\_Model schematic: depicts the shared encoder with classification and regression branches for ICU mortality and severity score predictions, along with the teacher–student knowledge distillation framework.}
    \label{fig:multitask_model}
\end{figure}

Figures~\ref{fig:mae_enc} and \ref{fig:mae_dec} illustrate our proposed MAE\_Encoder and MAE\_Decoder structures in detail. Initially, the input tokens 
($B \times 48 \times 7$) undergo a 5\% masking, and a [CLS] token is prepended to the sequence. This sequence is then linearly projected to 64 dimensions, with positional encoding added, before being processed by a 2-layer Transformer block configured with 
$d_{\text{model}} = 64$, a feed-forward network dimension of 256, 8 attention heads, and a dropout rate of 0.1. This same encoder architecture is later reused as the Multitask Encoder in both the teacher and student models, as shown in Figure~\ref{fig:multitask_model}.

\clearpage

The MAE\_Decoder reconstructs the masked tokens from the latent embedding (of size $B \times 48 \times 64$, excluding the CLS token) using a linear layer that converts the embedding back to 7 features per time step ($B \times 48 \times 7$). Once the teacher model is trained, its parameters are frozen, and it serves as a fixed guide for knowledge distillation to the student. In our multitask framework, the same encoder processes the $48 \times 7$ VIS input. For the classification task, the 64-dimensional CLS token is concatenated with full static features (51 dimensions after one-hot encoding) and fed to the classification layers, whereas for regression, the CLS token is concatenated with static features excluding 4 severity scores, yielding a 47-dimensional feature vector. The dropout rate for the Multitask Decoder is set to 0.2.

\section{MAE Pretraining and Teacher–Student Distillation}
\label{sec:method5}

\subsection{MAE Pretraining}
Let $\mathbf{X} \in \mathbb{R}^{B \times T \times 7}$ denote the VIS time series for a batch of $B$ patients, where $T=48$ represents the 48-hour monitoring period and 7 denotes the number of VIS features. To enable self-supervised learning, we randomly mask 5\% of the time steps or feature dimensions in $\mathbf{X}$, yielding a masked input $\mathbf{X}' = \mathrm{Mask}(\mathbf{X})$. This masked input is then processed by the MAE\_Encoder—a Transformer-based module—that encodes the partial information into latent representations. The corresponding MAE\_Decoder reconstructs the missing portions to generate $\hat{\mathbf{X}}$. The reconstruction loss is computed using the mean squared error (MSE) only over the masked positions, as defined in Equation~\eqref{eq:mae_loss}:

\begin{equation}
\mathcal{L}_{\mathrm{MAE}} =
\mathrm{MSE} \bigl(\hat{\mathbf{X}}_{\text{masked}}, \mathbf{X}_{\text{masked}} \bigr).
\label{eq:mae_loss}
\end{equation}

This objective forces the encoder to learn robust representations that capture the underlying structure of the VIS data despite missing or irregularly sampled values.

\subsection{Teacher Setup}
After completing MAE pretraining, the learned encoder is repurposed as the Teacher\_Encoder. In this phase, the pretrained encoder weights are loaded into a teacher model, which is then fixed—meaning no further gradient updates are applied during subsequent training on labeled data. The teacher model serves as a stable reference by providing fixed encodings or classification logits for knowledge distillation. It is important to note that the teacher does not undergo any additional multi-task supervision (i.e., no further cross-entropy or MSE loss on labeled data is applied), thereby preserving the robust representations it learned during the self-supervised phase. This frozen teacher model is then used to guide the training of the student model in later stages.

\subsection{Student and Distillation}
After the teacher model has been established, a student model with an identical architecture is trained on labeled data $(\mathbf{X}, y, s)$, where $y$ is a binary mortality label and $s$ represents auxiliary continuous targets (e.g., severity scores). The student model is designed to perform multitask learning by simultaneously addressing both classification and regression objectives. Its loss function, given in Equation~\eqref{eq:student_loss}, is a weighted combination of the two tasks:

\begin{equation}
\label{eq:student_loss}
\mathcal{L}_{\text{student}} =
\lambda_{\text{cls}} \, \mathrm{CE} \bigl( \mathbf{z}_S^{(\text{cls})}, y \bigr) +
\lambda_{\text{reg}} \, \mathrm{MSE} \bigl( \mathbf{z}_S^{(\text{reg})}, s \bigr),
\end{equation}
where $\mathbf{z}_S^{(\text{cls})}$ and $\mathbf{z}_S^{(\text{reg})}$ are the student’s outputs for classification and regression, respectively, and the hyperparameters $\lambda_{\text{cls}}$ and $\lambda_{\text{reg}}$ balance the two tasks.

In addition to this multitask objective, we incorporate knowledge distillation (KD) to further enhance the student model's performance. To achieve this, we compute the softmax probabilities for both teacher and student, as described in Equation~\eqref{eq:softmax}:

\begin{equation}
\label{eq:softmax}
p_T = \mathrm{softmax} \bigl( \mathbf{z}_T^{(\text{cls})} \bigr), \quad
p_S = \mathrm{softmax} \bigl( \mathbf{z}_S^{(\text{cls})} \bigr).
\end{equation}

The distillation loss, shown in Equation~\eqref{eq:kd_loss}, is defined as the squared Euclidean distance between these probability distributions:

\begin{equation}
\label{eq:kd_loss}
\mathcal{L}_{\text{KD}} =
\| p_S - p_T \|^2.
\end{equation}

Finally, the overall student loss function is augmented with the distillation term, as formalized in Equation~\eqref{eq:student_total_loss}, where $\lambda_{\text{KD}}$ controls the distillation strength:

\begin{equation}
\label{eq:student_total_loss}
\mathcal{L}_{\text{student}} \leftarrow
\mathcal{L}_{\text{student}} + \lambda_{\text{KD}} \, \mathcal{L}_{\text{KD}}.
\end{equation}

Because the teacher model remains frozen, it provides a stable reference that guides the student in learning robust classification boundaries, thereby improving generalization on noisy or incomplete ICU data. This integrated approach enables the student model to inherit the teacher’s distilled knowledge while concurrently learning task-specific features from the labeled dataset.

\subsection{Additional Parameter Explanations}
Throughout the paper, we use the following notations: $\mathbf{X}$ denotes the input VIS sequences of size $B \times 48 \times 7$, where $B$ represents the batch size. The binary mortality label is represented by $y$, and $s$ denotes the continuous severity score. The student’s classification logits and regression outputs are represented by $\mathbf{z}^{(\text{cls})}_S$ and $\mathbf{z}^{(\text{reg})}_S$, respectively, while the teacher and student classification probability distributions (obtained via softmax) are denoted by $\mathbf{p}_T$ and $\mathbf{p}_S$. The loss function incorporates weighting coefficients $\lambda_{\mathrm{cls}}$, $\lambda_{\mathrm{reg}}$, and $\lambda_{\mathrm{KD}}$ for the classification, regression, and distillation components, respectively.

In our implementation, we adopt an approximate 5\% masking ratio for the 48-hour VIS data. Although computer vision applications sometimes employ higher masking ratios (e.g., 30\%), a 5\% ratio strikes a balance by providing a sufficient challenge for the model while ensuring stable training. Excessive masking can destabilize early reconstruction, whereas insufficient masking may not fully leverage the model’s ability to learn robust representations.

The loss coefficients are chosen so that $\lambda_{\mathrm{cls}}=1.0$, thereby giving primary focus to the classification task through the cross-entropy loss, while $\lambda_{\mathrm{reg}}=0.1$ ensures that the regression loss (for severity scores) acts as a complementary signal. The distillation loss, which aligns the student’s classification output with that of the teacher, is weighted by $\lambda_{\mathrm{KD}}=0.05$.

Moreover, the teacher model, initialized from the pretrained MAE encoder, is not fine-tuned on the labeled data; it is set to \texttt{eval()} mode with all parameters frozen (i.e., \texttt{requires\_grad=False}). Only the student model’s encoder and its classification/regression heads are updated during training. We optimize using AdamW with a learning rate of $10^{-3}$, a weight decay of $10^{-4}$, and a batch size of 64. Each training stage—MAE pretraining, teacher setup, and student training—is run for up to 20–30 epochs, with early stopping employed if the validation AUROC fails to improve for several epochs. This training protocol has been found to yield stable convergence and improved generalization across both tasks.

\subsection{Overall Training Procedure} \label{sec:alg7_1}

Our overall pipeline proceeds in three distinct stages. First, we perform MAE pretraining on unlabeled VIS data, during which the encoder learns robust representations by reconstructing masked portions of the input (as quantified by the reconstruction loss in Equation~\eqref{eq:mae_loss}). Next, the pretrained encoder is loaded into a teacher model, whose parameters are then frozen; this teacher model provides fixed encodings or classification logits that serve as a stable reference for knowledge distillation. Finally, we train a student model—of the same architecture—on labeled data $(\mathbf{X}, y, s)$ using a composite loss function. This loss comprises the primary objectives of binary mortality classification and severity score regression (as defined in Equation~\eqref{eq:student_loss}), augmented by a knowledge distillation loss (see Equation~\eqref{eq:kd_loss}). The hyperparameter $\lambda_{\mathrm{KD}}$ controls the influence of the distillation term, ensuring that the student model gradually aligns its classification output with that of the teacher while simultaneously learning from the labeled data. 

Algorithm~\ref{alg:framework} shows the step-by-step process of this overall pipeline, detailing how the MAE pretraining, teacher setup, and student training stages are executed. This three-stage approach enables the student to inherit the robust representations developed during MAE pretraining and further refine them to achieve improved performance on both classification and regression tasks.

\begin{algorithm}[!ht]
\caption{Algorithm 4.5: Overall Pipeline (Right-Aligned Comments)}
\label{alg:framework}
\begin{algorithmic}[1]

\REQUIRE Labeled data $(\mathbf{X}, y, s)$; (optional) unlabeled data $\mathbf{X}_{\text{unlab}}$; mask ratio $\approx 0.05$
\Comment{\hfill Notation: $\mathbf{X}$ for inputs; $y$: mortality; $s$: severity.}

\STATE \textbf{Step 1: MAE Pretraining}
\Comment{\hfill (Line 1) Self-supervised reconstruction on unlabeled data.}

\STATE Initialize $\theta_{\text{MAE}}$
\Comment{\hfill (Line 2) $\theta_{\text{MAE}}$ are MAE parameters (encoder+decoder).}

\FOR{epoch $=1$ to $E_{\text{pre}}$}
   \FOR{mini-batch $\mathbf{X}_b$ in unlab\_loader}
       \STATE $\mathbf{X}_b' \leftarrow \mathrm{Mask}(\mathbf{X}_b)$
       \Comment{\hfill (Lines 5) Randomly mask $\approx5\%$ VIS positions.}
       \STATE $\mathbf{H} \leftarrow \mathrm{MAE\_Encoder}(\mathbf{X}_b'; \theta_{\text{MAE}})$
       \STATE $\hat{\mathbf{X}}_b \leftarrow \mathrm{MAE\_Decoder}(\mathbf{H}; \theta_{\text{MAE}})$
       \STATE $\mathcal{L}_{\text{MAE}} \leftarrow \mathrm{MSE}\bigl(\hat{\mathbf{X}}_{b,\text{masked}},\, \mathbf{X}_{b,\text{masked}}\bigr)$
       \Comment{\hfill (Lines 5--9) MAE objective on masked positions.}
       \STATE update $\theta_{\text{MAE}}$
   \ENDFOR
\ENDFOR

\STATE Save/freeze $\theta_{\text{MAE}}$ as \texttt{Teacher\_Encoder}
\Comment{\hfill (Line 12) End of MAE pretraining; teacher remains fixed.}

\STATE \textbf{Step 2: Teacher model setup (Frozen)}
\Comment{\hfill (Line 13) Load MAE encoder into teacher, set eval()}

\STATE Load $\theta_{\text{MAE}}$ into a teacher model
\Comment{\hfill (Line 14) No gradient updates on teacher.}

\STATE \textbf{Step 3: Student training with optional KD}
\Comment{\hfill (Line 15) Student learns from $(\mathbf{X}, y, s)$.}

\STATE Initialize $\theta_S$ (student parameters)
\Comment{\hfill (Line 16) Student net has classification/regression heads.}

\FOR{epoch $=1$ to $E_{\text{stud}}$}
   \FOR{$(\mathbf{X}_b, y_b, s_b)$ in train\_loader}
       \STATE $(\mathbf{z}^{(\text{cls})}_T, \_) \leftarrow \mathrm{Teacher}(\mathbf{X}_b)$
       \Comment{\hfill (Line 19) Teacher forward pass, no grad.}
       \STATE $(\mathbf{z}^{(\text{cls})}_S, \mathbf{z}^{(\text{reg})}_S) \leftarrow \mathrm{Student}(\mathbf{X}_b; \theta_S)$
       \Comment{\hfill (Lines 19--20) Student classification \& regression.}
       \STATE $\mathcal{L} \leftarrow \lambda_{\mathrm{cls}}\mathrm{CE}(\mathbf{z}^{(\text{cls})}_S,\,y_b)
         + \lambda_{\mathrm{reg}}\mathrm{MSE}(\mathbf{z}^{(\text{reg})}_S,\,s_b)$
       \Comment{\hfill Multi-task loss.}
       \IF{KD is enabled}
         \STATE $\mathbf{p}_T \leftarrow \mathrm{softmax}(\mathbf{z}^{(\text{cls})}_T)$
         \STATE $\mathbf{p}_S \leftarrow \mathrm{softmax}(\mathbf{z}^{(\text{cls})}_S)$
         \STATE $\mathcal{L}_{KD} \leftarrow \|\mathbf{p}_S - \mathbf{p}_T\|^2$
         \STATE $\mathcal{L} \leftarrow \mathcal{L} + \lambda_{KD}\,\mathcal{L}_{KD}$
         \Comment{\hfill (Line 26) Align student’s classification distribution with teacher’s.}
       \ENDIF
       \STATE update $\theta_S$
       \Comment{\hfill (Line 28) Gradient step on student.}
   \ENDFOR
   \STATE \text{(Stop if val AUROC plateaus for 5 epochs)}
   \Comment{\hfill (Line 30) Early stopping on validation.}
\ENDFOR

\RETURN final Student model

\end{algorithmic}
\end{algorithm}

\section{Results and Ablation Analysis}
\label{sec:results_ablation}
In this section, we present our multi-task Teacher--Student framework with MAE pretraining and analyze its ablation variants.

\subsection{Metric Definitions and Formulas}

To evaluate the performance of our binary classification model for ICU mortality prediction, we employ a comprehensive set of metrics that capture both the discriminative power and diagnostic utility of the model. Specifically, we use the area under the ROC curve (AUROC) to assess how well the model distinguishes between death and survival across various decision thresholds. In addition, we report the Positive Predictive Value (PPV) and Negative Predictive Value (NPV) to quantify the reliability of the model’s positive and negative predictions, respectively. Sensitivity (or recall) and specificity are calculated to measure the model's ability to correctly identify actual death and survival cases. Furthermore, we compute the Positive Likelihood Ratio (PLR) and Negative Likelihood Ratio (NLR) to evaluate the diagnostic strength of positive and negative predictions, along with overall accuracy (ACC) as a measure of the proportion of correctly classified instances. Table~\ref{tab:classification_metrics} provides a summary of these metrics and their formulas, offering a concise reference for the evaluation criteria employed in our study.

\renewcommand{\arraystretch}{2} % Adjust the row height
\begin{table}[ht]
    \centering
    \caption{Binary classification metrics and their formulas. Here, \(TP\) (True Positives) refers to correctly predicted positive cases (death), \(TN\) (True Negatives) refers to correctly predicted negative cases (survival), \(FP\) (False Positives) refers to incorrectly predicted positive cases (survival misclassified as death), and \(FN\) (False Negatives) refers to incorrectly predicted negative cases (death misclassified as survival).}
    \begin{tabular}{ll|ll}
        \toprule
        \textbf{Metric} & \textbf{Formula} & \textbf{Metric} & \textbf{Formula} \\
        \midrule
        AUROC (AUC) & \( \int_{-\infty}^{\infty} \mathrm{TPR}(t) \, d\mathrm{FPR}(t) \) & Accuracy (ACC) & \( \frac{TP + TN}{TP + TN + FP + FN} \) \\
        Positive Predictive Value (PPV) & \( \frac{TP}{TP + FP} \) & Negative Predictive Value (NPV) & \( \frac{TN}{TN + FN} \) \\
        Sensitivity (Recall) & \( \frac{TP}{TP + FN} \) & Specificity & \( \frac{TN}{TN + FP} \) \\
        Positive Likelihood Ratio (PLR) & \( \frac{\mathrm{Sensitivity}}{1 - \mathrm{Specificity}} \) & Negative Likelihood Ratio (NLR) & \( \frac{1 - \mathrm{Sensitivity}}{\mathrm{Specificity}} \) \\
        \bottomrule
    \end{tabular}
    \label{tab:classification_metrics}
\end{table}

\subsection{Ablation Study}

We conducted an extensive ablation study to assess the contributions of knowledge distillation (KD) and multi-task learning (MT) within our full Teacher–Student multi-task framework. Table~\ref{tab:ablation_lstm} presents a comparison of three configurations—our full model (\texttt{baseline} with kd=True, mt=True), a model without KD (\texttt{no\_kd}, kd=False, mt=True), and a model without the severity regression branch (\texttt{no\_mt}, kd=True, mt=False)—along with results from a prior LSTM-based approach \cite{Ning2023}. Notably, the LSTM study did not report certain metrics (e.g., Sensitivity, Specificity, Accuracy), which are indicated by “$\times$” in the table. Additionally, the $R^2$ column represents the coefficient of determination for our regression task, a metric not provided in the LSTM study.

The \texttt{baseline} configuration, which combines both KD and MT, achieves the highest AUROC (0.8223) and demonstrates favorable overall classification performance, with PPV, NPV, PLR, and NLR values of 0.4483, 0.9103, 3.16, and 0.38, respectively. This configuration also maintains a balanced Sensitivity (0.7033) and Specificity (0.7767), resulting in an Accuracy of 0.7616. The regression branch yields an $R^2$ of approximately 0.217, indicating that the model can account for a significant fraction of the variance in severity scores. These results underscore the synergistic benefits of integrating multi-task learning with knowledge distillation on top of MAE pretraining.

In contrast, the \texttt{no\_kd} configuration—where knowledge distillation is removed while retaining multi-task learning—shows a slight decrease in AUROC to 0.8172, although PPV increases to 0.4864 and PLR rises to 3.68. This suggests that, without the teacher’s guidance, the model adopts a more conservative yet precise approach in predicting mortality. However, this comes at the expense of a reduction in overall discrimination (AUROC) and a drop in Sensitivity (0.6631), despite a marginal improvement in $R^2$ (approximately 0.221). 

On the other hand, the \texttt{no\_mt} configuration, which eliminates the severity regression branch, exhibits a further decline in AUROC to 0.8084, with PPV and NPV at 0.4776 and 0.8972, respectively. Sensitivity decreases to 0.6352, although Specificity remains relatively high at 0.8208. The absence of auxiliary severity supervision in this configuration results in the loss of regression capability and compromises the robustness of feature extraction for classification.

Overall, the ablation study demonstrates that both knowledge distillation and multi-task learning contribute significantly to model performance. The KD mechanism strengthens classification boundaries—particularly improving Sensitivity and AUROC—while the MT approach provides essential auxiliary signals through severity regression, as evidenced by the enhanced $R^2$ values and more robust classification metrics. Consequently, the full \texttt{baseline} model, which integrates both KD and MT, offers the most balanced improvement across all evaluation metrics.

\renewcommand{\arraystretch}{1.2} % Consistent row height
\setlength{\tabcolsep}{6pt} % Balanced column spacing

\begin{table}[!ht]
    \centering
    \caption{Ablation results (with 95\% CI) on the test set for three configurations (\texttt{baseline}, \texttt{no\_kd}, \texttt{no\_mt}), plus an LSTM-based approach \cite{Ning2023} for partial comparison. Missing (unreported) metrics are marked ``\(\times\)''.}
    \label{tab:ablation_lstm}

    {\fontsize{10pt}{12pt}\selectfont % Ensure uniform font size

    \begin{tabular}{lcccccc}
        \toprule
        \textbf{Config} & \textbf{KD} & \textbf{MT} & R$^2$ & \textbf{AUROC (95\% CI)} & \textbf{PPV (95\% CI)} \\
        \midrule
        \textbf{baseline}  
        & \checkmark & \checkmark & 0.2171  
        & 0.8223 (0.7998--0.8434)  
        & 0.4483 (0.4070--0.4917) \\
        
        \textbf{no\_kd}  
        & --         & \checkmark & 0.2214  
        & 0.8172 (0.7946--0.8396)  
        & 0.4864 (0.4403--0.5303) \\
        
        \textbf{no\_mt}  
        & \checkmark & --         & -5.9291 
        & 0.8084 (0.7825--0.8324)  
        & 0.4776 (0.4315--0.5222) \\
        
        \textbf{LSTM \cite{Ning2023}}  
        & \(\times\) & \(\times\) & \(\times\) 
        & 0.74 (0.70--0.77)  
        & 0.39 (0.35--0.44) \\
        \bottomrule
    \end{tabular}

    \vspace{1em} % Add spacing before next table

    \begin{tabular}{lcccc}
        \toprule
        \textbf{Config} & \textbf{NPV (95\% CI)} & \textbf{PLR (95\% CI)} & \textbf{NLR (95\% CI)} & \textbf{ACC (95\% CI)} \\
        \midrule
        \textbf{baseline}  
        & 0.9103 (0.8959--0.9245)  
        & 3.1573 (2.8139--3.5333)  
        & 0.3821 (0.3263--0.4434)  
        & 0.7616 (0.7421--0.7806) \\

        \textbf{no\_kd}  
        & 0.9041 (0.8881--0.9187)  
        & 3.6830 (3.2217--4.2358)  
        & 0.4113 (0.3538--0.4733)  
        & 0.7873 (0.7679--0.8070) \\

        \textbf{no\_mt}  
        & 0.8972 (0.8819--0.9130)  
        & 3.5559 (3.1208--4.0695)  
        & 0.4445 (0.3867--0.5033)  
        & 0.7827 (0.7642--0.8012) \\

        \textbf{LSTM \cite{Ning2023}}  
        & 0.89 (0.87--0.91)  
        & 2.56 (2.25--2.91)  
        & 0.47 (0.40--0.55)  
        & \(\times\) \\
        \bottomrule
    \end{tabular}

    \vspace{1em} % Add spacing before next table

    \begin{tabular}{lccc}
        \toprule
        \textbf{Config} & \textbf{Sensitivity (95\% CI)} & \textbf{Specificity (95\% CI)} & \textbf{ACC (95\% CI)} \\
        \midrule
        \textbf{baseline}  
        & 0.7033 (0.6553--0.7446)  
        & 0.7767 (0.7543--0.7990)  
        & 0.7616 (0.7421--0.7806) \\

        \textbf{no\_kd}  
        & 0.6631 (0.6139--0.7086)  
        & 0.8193 (0.7988--0.8401)  
        & 0.7873 (0.7679--0.8070) \\

        \textbf{no\_mt}  
        & 0.6352 (0.5868--0.6807)  
        & 0.8208 (0.8001--0.8418)  
        & 0.7827 (0.7642--0.8012) \\

        \textbf{LSTM \cite{Ning2023}}  
        & \(\times\) & \(\times\) & \(\times\) \\
        \bottomrule
    \end{tabular}
    } % Close font size scope

\end{table}

\subsection{ROC Curves Analysis} 
\label{sec:roc_curves}

\begin{figure}[!ht]
\centering
\includegraphics[width=0.6\textwidth]{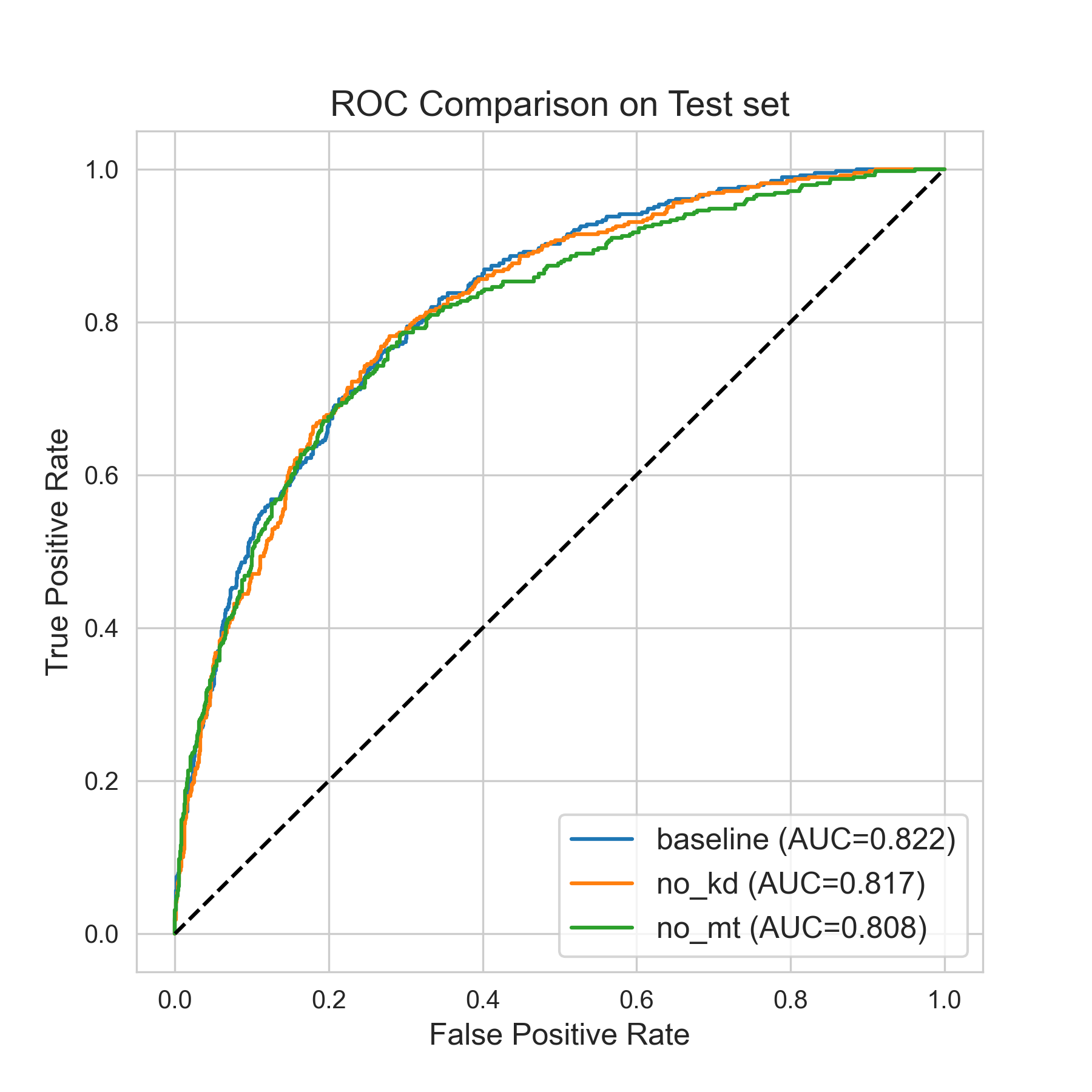}
\caption{ROC curves on the test set. \texttt{baseline} achieves the highest AUROC, followed by \texttt{no\_kd} and \texttt{no\_mt}, reflecting the synergy from knowledge distillation and multi-tasking.}
\label{fig:roccompare}
\end{figure}

Figure~\ref{fig:roccompare} shows the ROC curves of the three ablation configurations on the test set. The \texttt{baseline} model (kd=True, mt=True) achieves the highest AUROC (0.822), followed by \texttt{no\_kd} (0.817) and \texttt{no\_mt} (0.808). Notably, \texttt{baseline} maintains a better ROC curve across most thresholds, indicating stronger overall discrimination. Removing knowledge distillation (\texttt{no\_kd}) slightly reduces AUROC, while removing the severity regression branch (\texttt{no\_mt}) further decreases performance, illustrating the benefits of combining multi-task learning and distillation.

\subsection{Loss Curves and Convergence} \label{sec:loss_curves}

\begin{figure}[!ht]
    \centering
    \includegraphics[width=0.95\textwidth]{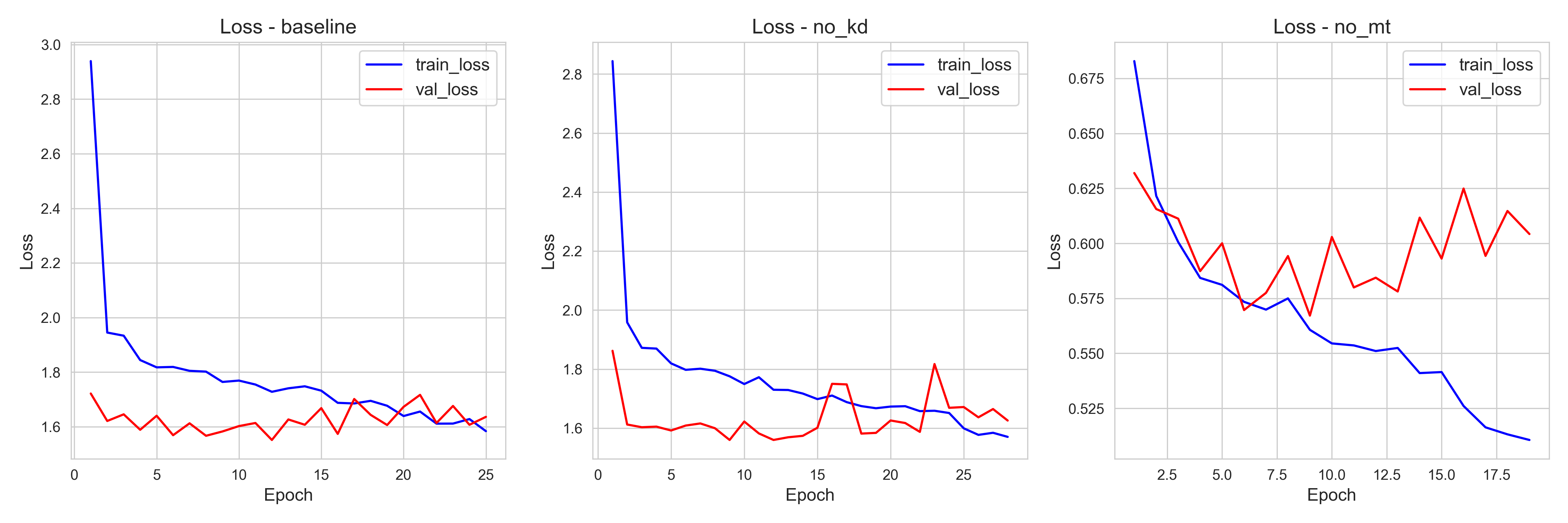}
    \caption{Training and validation loss curves for each ablation configuration. \texttt{baseline} (kd=True, mt=True) converges smoothly, while removing knowledge distillation (\texttt{no\_kd}) or multi-task learning (\texttt{no\_mt}) alters the loss trajectories and indicates reduced stability in later epochs.}
    \label{fig:loss_subplots}
\end{figure}

Figure~\ref{fig:loss_subplots} presents the training and validation loss trajectories for each ablation configuration. The \texttt{baseline} model (kd=True, mt=True) converges smoothly, suggesting that the synergy of knowledge distillation and multi-task objectives stabilizes training. In contrast, removing the severity regression branch (\texttt{no\_mt}) leads to a sharper spike in the validation loss during later epochs, indicative of reduced regularization. Meanwhile, \texttt{no\_kd} (no knowledge distillation) remains relatively steady but converges more slowly, underscoring the teacher model’s influence in guiding classification boundaries.

\subsection{SHAP Analysis for Interpretability} \label{sec:shap}

Table~\ref{tab:shap2} presents the top five static features ranked by mean absolute SHAP value, with \texttt{sofa\_score\_24h} demonstrating the highest impact (approximately 0.1467). This outcome aligns with clinical intuition, as an elevated SOFA score typically signifies more severe organ dysfunction and an increased likelihood of mortality. Other prominent features include \texttt{lods}, \texttt{marital\_status\_SINGLE}, \texttt{insurance\_Medicaid}, and \texttt{sapsii}, underscoring the combined relevance of both pathophysiologic and sociodemographic variables in sepsis outcomes.

\begin{table}[ht]
\centering
\caption{Top-5 static features by mean absolute SHAP. 
Socio-demographic factors appear alongside severity scores.}
\label{tab:shap2}
\begin{tabular}{lc}
\toprule
\textbf{Feature} & \textbf{mean\_abs\_shap}\\
\midrule
sofa\_score\_24h          & 0.1467 \\
lods                      & 0.0333 \\
marital\_status\_SINGLE   & 0.0307 \\
insurance\_Medicaid       & 0.0231 \\
sapsii                    & 0.0199 \\
\bottomrule
\end{tabular}
\end{table}

\begin{figure}[!ht]
\centering
\includegraphics[width=0.48\textwidth]{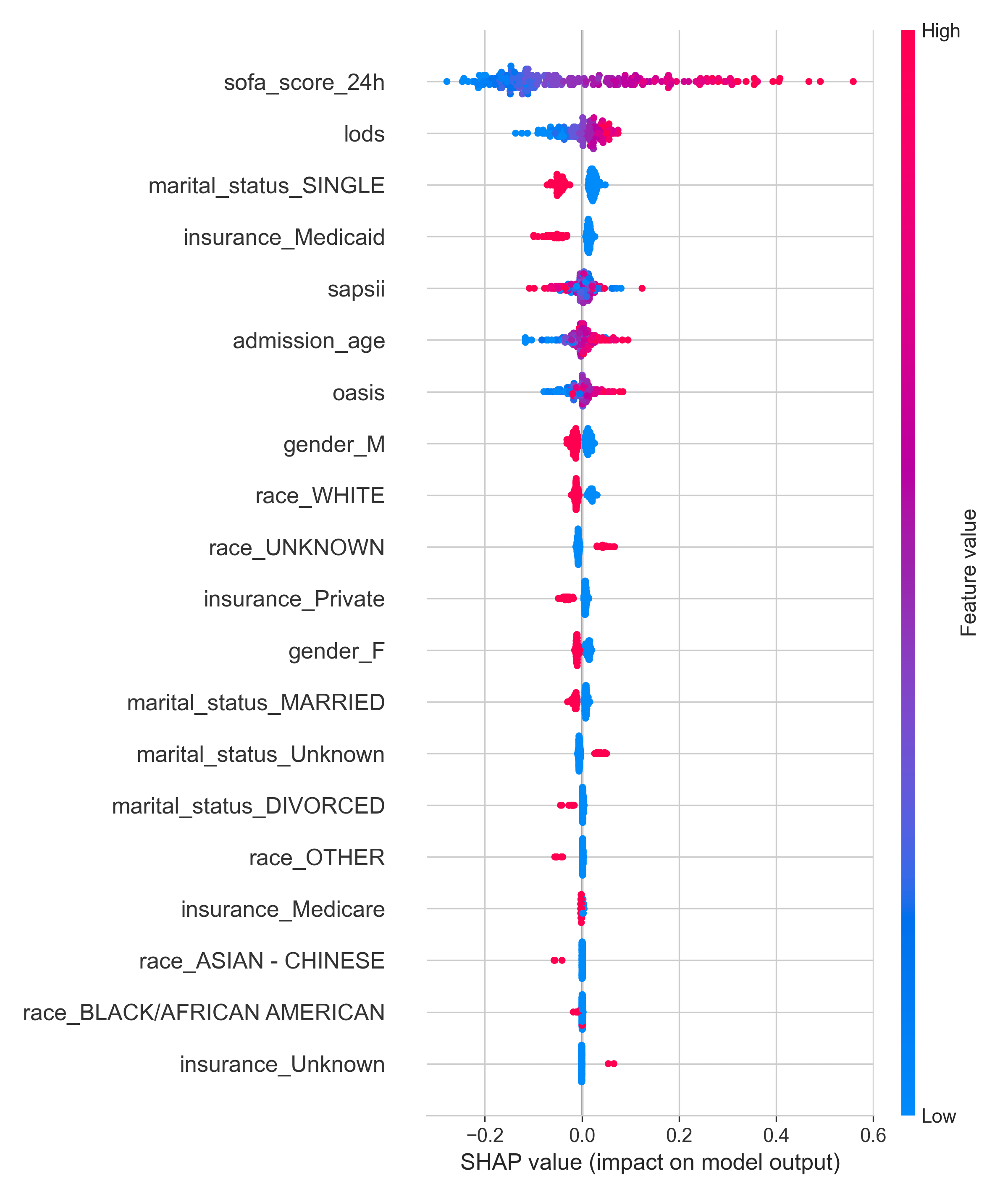}
\caption{SHAP summary plot illustrating the distribution of feature contributions to the model’s predictions.}
\label{fig:shapsummary}
\end{figure}

Figure~\ref{fig:shapsummary} illustrates the distribution of SHAP values, revealing that \texttt{sofa\_score\_24h} displays the broadest range and hence wields substantial influence over the model’s predictions. Likewise, the prominence of \texttt{lods} is consistent with its well-established role in quantifying organ dysfunction, which is a critical determinant of sepsis severity and subsequent mortality risk~\cite{Singer2016}. These findings reinforce the understanding that multiorgan failure is a principal driver of sepsis-related mortality in the ICU.

Notably, single marital status and Medicaid insurance emerge as influential factors, indicating that social support systems and healthcare access can significantly affect patient outcomes. This observation aligns with the work of Seymour et al.~\cite{Seymour2010} and Mitchell et al.~\cite{Mitchell2021}, who found that patients lacking robust social or financial support are more susceptible to delayed interventions and poorer prognoses. Even when organ dysfunction levels are similar, individuals with limited resources may experience worse outcomes, highlighting the nuanced interplay between physiological and socioeconomic determinants of sepsis mortality.

Although \texttt{sapsii} appears slightly lower on the SHAP importance list, it remains an integral severity scoring system that integrates various physiological and laboratory measurements. While partially overlapping with SOFA, SAPSII contributes additional insights into disease progression, bolstering the overall interpretability and accuracy of the model’s predictions.

Overall, these results illustrate that organ dysfunction indicators (e.g., SOFA, LODS) remain core drivers of ICU mortality prediction, yet demographic and socioeconomic factors (e.g., marital status, insurance type) can critically modulate risk. Patients exhibiting equivalent organ failure may nevertheless face higher mortality if they lack social support or adequate healthcare coverage. Consequently, the SHAP analysis underscores the importance of incorporating both clinical severity and socioeconomic variables in sepsis prognostication, thereby enabling a more holistic and equitable approach to identifying high-risk patients and informing timely, targeted interventions.

\section{Discussion} \label{sec:discussion}

Table~\ref{tab:ablation_lstm} includes the prior LSTM study~\cite{Ning2023} for comparison, which reported an AUROC of approximately 0.74 (95\% CI: 0.70--0.77), PPV = 0.39 (0.35--0.44), NPV = 0.89 (0.87--0.91), PLR = 2.56 (2.25--2.91), and NLR = 0.47 (0.40--0.55), but did not provide Sensitivity, Specificity, or Accuracy. In contrast, our \texttt{baseline} model (incorporating MAE pretraining, knowledge distillation, and multi-task learning) achieves an AUROC of about 0.82, with higher PPV (0.45 vs.\ 0.39), NPV (0.91 vs.\ 0.89), and PLR (3.16 vs.\ 2.56), alongside a lower NLR (0.38 vs.\ 0.47). These improvements across multiple metrics suggest that MAE pretraining effectively addresses missing or irregular VIS data by randomly masking and reconstructing time-series signals, knowledge distillation provides robust classification boundaries while balancing model size and accuracy, and multi-task learning with severity regression adds auxiliary signals that mitigate overfitting and enrich the learned representations.

A closer look at the LSTM-based approach of Ning et al.~\cite{Ning2023} helps clarify why our framework outperforms their model. In ICU settings, medication and physiological indicators often suffer from discontinuity and missing values, and the LSTM models in the comparative studies rely on the method proposed by Thorsen Meyer et al.~\cite{ThorsenMeyer2020} for processing dynamic time series. While this method offers a structured way to handle irregularly sampled data, it may not fully capture the latent patterns or relationships present in highly variable ICU data. In contrast, our MAE pretraining phase leverages random masking of the VIS sequence, compelling the network to reconstruct missing information and learn more noise-resistant embeddings. This approach reduces the likelihood of training instability or overfitting, resulting in features that are comprehensive and robust. Consequently, when these enriched features are fed into subsequent classification and regression tasks, the overall performance benefits from their resilience to data sparsity and fluctuations, thereby exceeding the capabilities of traditional LSTM methods.

\subsection{Guidance from Teacher–Student Knowledge Distillation} 
Traditional LSTM models are typically single architectures that must handle multiple feature dimensions, often leading to unstable training or insufficient generalization when faced with high-dimensional, complex dose sequences. By contrast, our approach leverages a “teacher model” obtained from MAE pretraining, which remains fixed during subsequent training. The student model inherits crucial classification boundaries and the “soft-label” distribution from the teacher through knowledge distillation. Empirically, this process maintains a higher AUROC and more stable Sensitivity, effectively reducing overfitting and improving mortality recognition compared to a direct training strategy without distillation.

\subsection{Collaborative Reinforcement of Data Features by Multi-Task Learning (MTL)} 
In the comparative LSTM study~\cite{Ning2023}, only a single mortality classification objective was employed, offering no direct supervision for organ function or severity scores (e.g., SOFA). In our approach, however, the model simultaneously predicts mortality and performs severity regression, thereby capturing the dynamic interplay between VIS changes and organ function. By learning both tasks concurrently, the shared layers receive more interrelated gradient information, forming richer internal representations. This advantage is reflected in our experimental results, where enabling multi-task learning boosts AUROC, NLR, and NPV, while the regression branch achieves an $R^2$ of over 0.2. These findings show that including severity regression provides valuable complementary supervision rather than conflicting with mortality prediction.

\subsection{More Flexible Handling of Irregular and Missing Data} 
ICU data, including VIS and other measurements, frequently contain missing values and exhibit extreme imbalance. The LSTM methods in the comparative studies rely on the strategy proposed by Thorsen Meyer et al.~\cite{ThorsenMeyer2020}, which involves discarding patient data that do not align with predefined time points (e.g., 1–48 h). This procedure risks losing informative samples and potential nuances in the data. In contrast, our MAE pretraining phase simulates random missingness through masking, enabling the model to learn robust inferences even with incomplete information. Furthermore, multi-task learning and knowledge distillation help focus the model on key features, thereby enhancing generalizability and mitigating the adverse effects of missing data.

\subsection{Demographic Information and Severity Scores as Features} 
A distinctive aspect of our framework is the integration of demographic information and various severity scores alongside the student model’s classification module. This design allows the network to capture differences in demographic distributions while concurrently leveraging the impact of severity scores on patient mortality. Consequently, the model gains a broader perspective on the factors influencing outcomes, encompassing both the clinical severity of organ dysfunction and the patient’s sociodemographic context.

\subsection{Experimental Results and Clinical Implications} 
In comparison to the LSTM approach, which achieved an AUROC of approximately 0.74, our proposed method attains an AUROC of around 0.82. Additionally, key metrics such as PPV, NPV, PLR, and NLR all show marked improvements, indicating that our model can more accurately identify high-risk patients and reduce misclassifications in a similar ICU setting. When combined with severity regression, the model further assists clinicians in understanding a patient’s condition trajectory, facilitating timely interventions and enabling more personalized treatment strategies.

\section{Study Limitations and Future Work}
Although this study achieved notable predictive performance and interpretability on the MIMIC-IV database, several directions warrant further exploration and refinement. First, despite MIMIC-IV’s breadth, it remains a single database. Previous research has highlighted that models trained and tested on the same source may lack external consistency. Accordingly, testing or retraining on other independent datasets would confirm whether the model maintains accuracy and robustness across varying healthcare settings, time frames, and patient populations.

In real clinical workflows, seamless integration into ICU information systems is essential for practical adoption. To achieve this, the model must provide real-time risk alerts and interpretable outputs (e.g., SHAP visualizations) without disrupting standard diagnostic and therapeutic procedures. Additionally, high-frequency data synchronization is critical, ensuring the model receives up-to-date electronic medical records, medication dosages, and monitoring indicators. From a usability standpoint, computing resource demands, latency, and interoperability with hospital systems must also be considered, particularly during peak care periods or emergency scenarios.

Beyond the current focus on 48-hour VIS dynamics and static characteristics (including severity scores and socioeconomic indicators), there exists a wide range of potentially valuable information in the ICU. Multimodal data, such as continuous physiological streams (heart rate, blood pressure, respiratory rate) and temporal laboratory indicators (blood gases, inflammatory markers), may offer deeper insights. Unstructured data sources, such as pathology images or free-text clinical notes, could further enhance the model’s understanding of sepsis progression. Incorporating these data streams within a similar MAE or multitask-distillation framework could yield even stronger predictive performance.

While retrospective validation is a common practice, prospective clinical trials are vital for evaluating the model’s utility under real-world constraints. Such trials would test performance amid fluctuating workloads, variable data quality, and varying degrees of physician adherence to model recommendations. Ultimately, this study demonstrates the potential of combining a multitask Teacher–Student architecture with MAE pretraining for early sepsis prognosis. However, extensive external validation and prospective research in more diverse populations and clinical contexts will be necessary for large-scale clinical implementation. Equally important is ensuring the model’s deeper integration with ICU processes to support timely interventions and improve patient outcomes.

\section{Conclusion}

This study conducted an in-depth modeling of vasoactive–inotropic score (VIS) dynamics in septic patients over a 48-hour period, integrating Masked Autoencoder (MAE) pretraining, Teacher–Student knowledge distillation, and Multi-Task Learning (MTL) on MIMIC-IV data. Our final model achieved notably strong performance on the test set, with an AUROC of approximately 0.8223, PPV of 0.4483, NPV of 0.9103, PLR of 3.1573, NLR of 0.3821, Sensitivity of 0.7033, Specificity of 0.7767, and Accuracy of 0.7616. Compared to traditional LSTM-based methods (which yielded an AUROC around 0.74), our approach not only improved the identification of high-risk patients (e.g., higher PPV, PLR) but also enhanced the reliable detection of surviving patients (higher NPV, Specificity) while maintaining a more balanced overall classification (Accuracy of about 0.76). Additionally, our multitask framework demonstrated a regression capability for severity scores, reflected in an $R^2$ exceeding 0.2, suggesting complementary support for assessing organ failure trends in clinical practice.

Mechanistically, MAE pretraining capitalizes on missing and irregular time-series data by randomly masking inputs and learning robust representations. Teacher–Student knowledge distillation further refines classification boundaries by transferring latent information from the teacher model to the student, preserving accuracy while reducing the risk of overfitting. Meanwhile, MTL leverages the shared objective of mortality classification and severity regression, leading to more comprehensive feature extraction and improved robustness to incomplete data. This synergy of MAE, knowledge distillation, and MTL ultimately strengthens both classification and regression performance, even in challenging ICU environments.

In practical clinical scenarios, the proposed model can be embedded into ICU information systems to track patients’ medications and vital indicators in real time, offering clinicians an early warning of mortality risk alongside a predictive severity score. Such insights facilitate earlier interventions, more efficient resource allocation, and improved patient prognosis. Furthermore, interpretability tools like SHAP provide transparency into the model’s decision logic, enabling healthcare providers to understand how clinical severity scores and sociodemographic factors shape patient outcomes. Nonetheless, additional validation and refinement in external databases and prospective trials remain essential for ensuring the model’s applicability across diverse populations and settings. With these steps, the proposed framework holds promise for enhancing ICU decision-making, optimizing sepsis management, and ultimately improving patient survival.

\section*{Acknowledgments}
The authors would like to thank all contributors to MIMIC-IV for providing the publicly accessible electronic health record (EHR) database.

\end{document}